\documentclass[conference]{IEEEtran}
\IEEEoverridecommandlockouts

\ifCLASSINFOpdf
\else
\fi
\usepackage{amsmath}
\usepackage{amsfonts}
\usepackage{multirow}
\usepackage{graphicx} 
\usepackage{caption}
\usepackage{subcaption}

\usepackage{algorithm}
\usepackage{algorithmic}
\usepackage{url}

\DeclareMathOperator*{\argmax}{argmax}
\DeclareMathOperator*{\argmin}{argmin}

\newcommand{\etal}{et al.}  
\graphicspath{ {./images/} }

\begin{document}
%
\title{Learning a metric for class-conditional KNN}

\author{\IEEEauthorblockN{Daniel Jiwoong Im\thanks{D.J.~Im performed
      the work while at the University of Guelph.}}
\IEEEauthorblockA{
Janelia Research Campus, HHMI\\
Email: imd@janelia.hhmi.org}
\and
\IEEEauthorblockN{Graham W. Taylor}
\IEEEauthorblockA{School of Engineering\\
University of Guelph\\
Guelph, Canada\\
Email: gwtaylor@uoguelph.ca}}

%


\maketitle
\vspace{-8mm}
\begin{abstract}
  Na\"ive Bayes Nearest Neighbour (NBNN) is a simple and effective
  framework which addresses many of the pitfalls of K-Nearest
  Neighbour (KNN) classification. It has yielded competitive results
  on several computer vision benchmarks. Its central tenet is that
  during NN search, a query is not compared to every example in a
  database, ignoring class information. Instead, NN searches are
  performed within each class, generating a score per class.  A key
  problem with NN techniques, including NBNN, is that they fail when
  the data representation does not capture perceptual
  (e.g.~class-based) similarity.  NBNN circumvents this by using
  independent engineered descriptors (e.g.~SIFT). To extend its
  applicability outside of image-based domains, we propose to
  \emph{learn} a metric which captures perceptual similarity. Similar
  to how Neighbourhood Components Analysis optimizes a
  differentiable form of KNN classification, we propose
  ``Class Conditional'' metric learning (CCML), which optimizes a soft
  form of the NBNN selection rule. Typical metric learning algorithms
  learn either a global or local metric.  However, our proposed method
  can be adjusted to a particular level of locality by tuning a single
  parameter.  An empirical evaluation on classification and retrieval
  tasks demonstrates that our proposed method clearly outperforms
  existing learned distance metrics across a variety of image and
  non-image datasets.
\end{abstract}


%
\IEEEpeerreviewmaketitle

\section{Introduction}
Nearest Neighbour (NN) techniques for image classification have been
overshadowed by parametric methods such as the bag-of-words method
\cite{Csurka04visualcategorization} combined with the spatial pyramid
match kernel \cite{lazebnik2006beyond}, and more recently,
convolutional nets \cite{Krizhevsky:2012,szegedy2014going}. However,
Nearest Neighbour methods have several appealing properties: they are simple, they can handle a large
number of classes that are not necessarily defined \emph{a priori},
with no free parameters they cannot overfit, and they require no separate
training phase.

An issue central to NN methods is the distance used to find
neighbours. It is well known that Euclidean distance in pixel space is
flawed. It is computationally expensive to compute on images of
non-trivial size and it tends not to capture perceptual similarity,
meaning that examples that humans perceive to be similar may not lie
close together in Euclidean space. Modern NN methods
\cite{Boiman2008,mccann2012local} rely on combinations of feature
descriptors such as SIFT \cite{lowe2004distinctive}, luminance, color,
and shape descriptors \cite{mori2005efficient}. The downside of using
such descriptors are that they are domain-specific and require
engineering. They also tend to produce high-dimensional feature
spaces. In the bag-of-words/spatial pyramid paradigm, descriptors are
quantized. However, Boiman \etal~argue in a seminal paper
\cite{Boiman2008} that quantization, though significantly reducing
dimensionality, leads to a severe degradation in the discriminative
power of descriptors. Parametric methods can compensate by learning,
but non-parametric methods are considerably weakened having
essentially no way to correct.

A second issue hindering NN methods 
is the use of ``image-to-image'' rather than ``image-to-class''
distance. The former sees a query image compared to every other image
in a database, selecting the $k$ nearest points. The latter sees a
query image compared to each \emph{class} in turn, selecting the $k$
nearest points in each class. This idea is key to a simple algorithm
named Na\"ive-Bayes Nearest Neighbours (NBNN). The descriptors of the
query image are strongly assumed to be independent by the model;
however, the algorithm works well in practice and has seen several
extensions which rival the state-of-the-art in object recognition
\cite{behmo2010towards,mccann2012local}.

Returning to the central issue of distance in NN algorithms, an
alternative to using descriptors is to \emph{learn} a distance measure
directly from data, exploiting distance information that is
intrinsically available, for example, through class
labels.
Metric learning methods aim to learn a
similarity measure that is both tractable to compute and perceptually
coherent (i.e. observations that are perceptually similar will have a
high measurable similarity).

Many distance metric learning methods are related to techniques for
dimensionality reduction\cite{Weston2008, Bian2014, Xu2014}. Simple methods like Locally Linear Embedding (LLE)
\cite{Saul2003} and Stochastic Neighbor Embedding (SNE)
\cite{Hinton2002} depend on meaningful similarity in the input
space. These approaches are not suitable for images unless they are
perfectly registered and highly similar. Laplacian Eigenmaps
\cite{Belkin2003} and other spectral methods do not require a
meaningful metric in the input space, but they cannot cope with
out-of-sample data because they do not learn a
mapping\footnote{Extensions of these techniques to cope with the out-of-sample problem
  have been proposed (c.f.~\cite{bengio2004out,carreira2007laplacian,kim2009covariance}).}. This limits
their applicability to retrieval.

A subset of approaches learn a function (mapping) from
high-dimensional (i.e. pixel) space to low-dimensional ``feature''
space such that perceptually similar observations are mapped to nearby
points on a manifold.  Neighbourhood Components Analysis (NCA)
\cite{Goldberger} proposes a solution where the transformation
from input to feature space is linear and the distance metric is
Euclidean.
NCA learns a distance metric which optimizes a smooth approximation of
$k$-NN classification. While it is optimized for $k=1$, using $k>1$
tends to work better in practice, and a recent extension
\cite{tarlow2013stochastic} derives a tractable way to optimize for
$k>1$ neighbours. NCA can be extended to the nonlinear case
\cite{Salakhutdinov2007} by introducing intermediate representations
and applying back-propagation.  NCA is fundamentally
``image-to-image'' and training it requires a normalization over every
pair of points.

In this paper, we introduce a smooth form of ``image-to-class''-based
Nearest Neighbour in which, in the spirit of NCA, distances are
computed via a discriminatively learned embedding. In contrast to NCA,
our algorithm makes local rather than global changes in its updates:
nearest neighbours in the same class are pulled towards a point, while
nearest neighbours in other classes are repelled. Far away points are
unaltered. We show that the ``image-to-class'' paradigm works not just
with engineered descriptors, but with learned distance metrics.
Furthermore, learning the distance metric according to an
``image-to-class'' criterion is effective.

\section{Background}
Before developing our technique, we first give a brief overview of Na\"ive Bayes
Nearest Neighbour \cite{Boiman2008} and then Neighbourhood Components Analysis \cite{Goldberger}.


\subsection{Na\"ive Bayes Nearest Neighbour}

The goal of NBNN \cite{Boiman2008} is to classify a query image.
Each such query image $Q$ is classified as belonging to
a class $\hat{C}$ according to a \emph{maximum a posteriori}
estimation rule

\begin{equation}
    \hat{C} = \argmax_C p(C|Q) ~.\
\end{equation}

Assuming a uniform class prior and applying Bayes' rule,

\begin{equation}
\hat{C} = \argmax_C \log p(Q|C) ~.\
\end{equation}

In Na\"ive-Bayes Nearest Neighbours and its variants, a set of
descriptors $\{d_1, \ldots, d_n\}$ is extracted from each
image. Assuming independence of the descriptors $d_i$, given the class

\begin{align}
    \hat{C} & \! =\! \argmax_C \!\! \left[ \log \prod_{i=1}^n p(d_i|C) \right]
             \!\! = \! \argmax_C \!\! \left[ \sum_{i=1}^n \log p(d_i|C) \right].\
     \end{align}

Approximating $p(d_i|C)$ by a Parzen window estimator with kernel
$\mathcal{K}$ leads to

\begin{equation}
\hat{p}(d_i|C) = \frac{1}{L} \sum_{j=1}^L \mathcal{K}\left( d_i -
  d_j^C \right) ~,\
\label{parzenapprox}
\end{equation}

\noindent where $L$ is the total number of descriptors over all
images from class $C$ in the training set and $j$ indexes descriptors
in class $C$. This can be further approximated by
only using the $k$ nearest neighbour descriptors in class $C$,
\begin{equation}
p(d_i|C) \approx \hat{p}_k(d_i|C) = \frac{1}{L} \sum_{j=1}^k \mathcal{K}\left( d_i -
  d_{NN_j}^C \right) ~,\
\label{knnkernel}
\end{equation}
\noindent


\noindent where $d_{NN_j}^C$ is the
$j$\textsuperscript{th} nearest neighbour of $d_i$ in class $C$. By choosing a Gaussian
kernel for $\mathcal{K}$ and substituting this into
Eq.~\ref{knnkernel}, the classification rule becomes

\begin{align}
\hat{C} & = \argmax_C \left[ \sum_{i=1}^n \log \frac{1}{L} \sum_{j=1}^k \exp^{-\frac{1}{2
      \sigma^2} \|d_i -NN_j^C(d_i)\|^2} \right] ~.
\end{align}

By setting $k=1$, there is no longer a dependence on the kernel width
$\sigma$ and $\log p(d_i|C)$ has a very simple form:

\begin{align}
\hat{C} = \argmin_C \left[
  \sum_{i=1}^n \|d_i - \text{NN}^C(d_i)\|^2 \right] ~,\
  \label{eqn:nbnneqn}
\end{align}

\noindent where $\text{NN}^C(d_i)$ is the nearest neighbour to $d_i$
in class $C$. This is the setting used in
\cite{Boiman2008} and \cite{mccann2012local}.

In our approach, we propose to avoid the task of extracting
descriptors and instead learn a functional mapping $z = f(x)$, where
$x$ is a vector of the pixels of image $Q$ and $z$ is a vector
representation in some feature space. Nearest neighbours are then
found in this feature space. Before discussing this adaptation, we
review Neighbourhood Components Analysis \cite{Goldberger}, a method
which learns a functional mapping by optimizing a soft form of the KNN
classification objective. Although NBNN is motivated from the
standpoint of image classification, the idea of ``image-to-class''
comparison is broadly applicable to non-image modalities. Indeed, in
our experiments, we consider non-image data. Additionally, once we have
dropped the dependence on descriptors in favour of a single feature
vector, we no longer rely on the Na\"ive-Bayes assumption. Therefore,
in the remainder of the paper we will use the term
``class-conditional'' to refer to the ``image-to-class'' variant of
KNN implied by NBNN.


\subsection{Neighbourhood Components Analysis}\label{sec:NCA}

NCA aims to optimize a distance metric under the objective of 1-NN
classification accuracy. The form of the mapping $f$ is not
fundamental to the technique, only that it is differentiable and
therefore trainable by gradient-based methods. To simplify our
presentation and without loss of generality, we follow
\cite{Goldberger} and adopt a linear mapping $z = Ax$. The distance
between two vectors $x$ and $x'$ is then defined as

\begin{equation}
g_A \left(x, x'\right) = \left(x - x'\right)^T A^T A \left(x -
  x'\right).
\end{equation}

This can be interpreted as first projecting points $x \in
\mathbb{R}^D$ into $P$-dimensional space using $A \in \mathbb{R}^{P
  \times D}$ and then computing Euclidean distances in the
$P$-dimensional space.

Optimizing $A$ with respect to the KNN classification criterion is not
differentiable because the objective is not a smooth function of $A$.
For example, making a small change to $A$ may change the nearest-neighbour
assignment, leading to a large change in the objective, whereas a
large change to $A$ may have negligible effect on the objective if it
does not impact neighbour selection. Thus, NCA adopts a
probabilistic view, in which a point $i$ selects a point $j$ to be its
neighbour with a probability that is a function of their difference in
feature space. The learning objective becomes the expected
accuracy of a 1-NN classifier under the probability distribution,

\begin{equation}
L(A) = \sum_i \sum_{j \neq i} p_i(j)[C_i = C_j],
\end{equation}

\noindent where $p_i(j) \propto \exp ( - \|Ax_i - Ax_j\|_2^2 )$ is the
probability that $i$ selects $j$ as its neighbour.

\section{Class Conditional Metric Learning}

Note that NCA takes essentially the ``image-to-image'' view, in which
each point is compared to every other point based on its distance in
projected space. Computing $p_i(j)$ is quadratic in the number of
training points which severely limits its applicability to large
datasets. In practice, NCA is often trained approximately using
``mini-batch'' gradient descent \cite{Salakhutdinov2007}.
%
Inspired by NBNN, we propose an alternative objective to distance
metric learning having two key properties: (i) nearest-neighbour
finding is class-conditional; and (ii) only the $k$-nearest neighbours
in each class contribute to the class decision. Let the probability of
assigning point $i$ to class $C$ be a normalized function of the distances to its
nearest neighbours \emph{in that class}:
%
\begin{equation}
    p_{i}^C = \frac{\exp\left(-\frac{1}{k}\sum_{j=1}^k || Ax_i - \text{NN}^C_j(Ax_i) ||^2\right)}
         {\sum_{C'} \exp\left(-\frac{1}{k}\sum_{j=1}^k || Ax_i -
             \text{NN}^{C'}_j(Ax_i) ||^2\right)} ~,\
    \label{eqn:org_cne}
\end{equation}

\noindent where $\text{NN}^C_j(Ax_i)$ is the $j$\textsuperscript{th} nearest
neighbour, from class C, of the projection of $x_i$.

\begin{algorithm}[t]
    \caption{Mini-batch stochastic gradient descent training of Class Conditional Metric Learning (CCML). }
    \label{algo:ccml}
    \begin{algorithmic}
        \FOR {$m$ iterations}
        \STATE Sample a
mini-batch from the training data
        \STATE Compute the probability $p_{i}^{C_i}$ $\forall$ $C_i$ using Eq.~\ref{eqn:org_cne}
        \STATE Compute the expectation in Eq.~\ref{eqn:expectation}
        \STATE Update the parameters $A$ proportional to Eq.~\ref{eqn:grad_ccml}
        \ENDFOR
    \end{algorithmic}
\end{algorithm}
\vspace{-1mm}

%
Similar to NCA, our objective is to maximize the expected accuracy of
classification under this objective:
%

\begin{align}
E(A) &= \sum_i \sum_{C} p_{i}^C \left[ C = C_i \right] = \sum_i
p_{i}^{C_i} ~,\
\label{eqn:expectation}
\end{align}

\noindent where $C_i$ is the class associated with point $x_i$.
At each update, each data point will be attracted toward the $k$
nearest data points that share its class. This is equivalent to
minimizing the mean distance between point $i$ and its $k$ nearest
neighbours in the same class. This will encourage similar points in
the same class to cluster. The denominator in Eq.~\ref{eqn:org_cne}
will locally repel nearby points not in the same class as $i$.

Differentiating the expectation \ref{eqn:expectation} with respect to
the transformation matrix $A$ gives

\begin{multline}
    \label{eqn:grad_ccml}
    \frac{\partial E}{\partial A} = - \frac{1}{k} 2 \sum_i p_{i}^{C_i} (\beta^{C_i}-\\
     \frac{\sum_{C'} \beta^{C'} \exp{(-\frac{1}{k}\sum_{j=1}^k || Ax_i - \text{NN}^{C'}_j(Ax_i) ||^2})}
    {\sum_{C'}  \exp{(-\frac{1}{k}\sum_{j=1}^k || Ax_i -
        \text{NN}^{C'}_j(Ax_i) ||^2})}) ~,\
\end{multline}

\noindent where

\begin{equation}
\beta^{C_i} = \sum_{j=1}^k(Ax_i- \text{NN}^{C_i}_j(Ax_i))x_i^T .\
\end{equation}

Our method, which we call Class Conditional Metric Learning (CCML) can be
trained by mini-batch stochastic gradient descent\footnote{In
  practice, we train with stochastic mini-batches, only performing the
  nearest neighbour search over the mini-batch.}, using
Eq.~\ref{eqn:grad_ccml} to compute the gradient needed for the update
to $A$.

Strictly speaking, $\text{NN}(\cdot)$ is a non-differentiable function, because its
  derivative is not defined at the boundaries between nearest neighbour assignments.
  However, in practice, we did not experience problems during
  optimization\footnote{We used the Theano library, specifically its auto-differentiation capabilities. We verified its stability of taking the gradient of $\text{NN}(\cdot)$ with the Theano development team (personal communication).}.
  This is because most of the points lie on the interior of the nearest neighbour
  boundaries and the use of mini-batch SGD compensates
  when the gradients are averaged.
  A theoretically sound way to overcome this issue is by running SGD
  and ignoring the updates when we encounter a
  non-differentiable point \cite{bottou1998}.
%
%

  \subsection{Test time decision rule}\label{sec:ccknn}
At test time, we project each query, $z = Ax$ and apply the following decision rule, which we call
Class Conditional KNN (CCKNN):
%

\begin{align}
\hat{C} = \argmin_C \left[
  \sum_{j=1}^k ||z - \text{NN}^C_j(z)||^2 \right].
  \label{eqn:ccknn}
\end{align}
Eq.~\ref{eqn:ccknn} has a similar form to NBNN (Eq.~\ref{eqn:nbnneqn}),
except that for each class, we sum over the $k$
nearest neighbours of the projection of the query, $z$. This is
consistent with Eq.~\ref{eqn:org_cne}, the probability of a query
point $i$ being assigned to class $C$, $p_{i}^{C} \propto
\exp{(-\sum_{j=1}^k ||z_i - \text{NN}^{C}_j(z_i)||^2 )}$. The class
with the highest probability under the ``soft'' decision rule is exactly
the class selected by Eq.~\ref{eqn:ccknn}. We are explicit in
differentiating the metric learning algorithm (CCML) from the nearest neighbour
decision rule (CCKNN) so that we can evaluate their respective
contributions in Section \ref{experiments}.
Pseudo-code for training and testing our proposed model using
CCML and CCKNN is shown in
Algorithm~\ref{algo:ccml} and~\ref{algo:ccknn}, respectively.
\begin{figure*}[ht]
\centering
    \begin{minipage}[b]{0.9\linewidth}
    \includegraphics[width=1\textwidth]{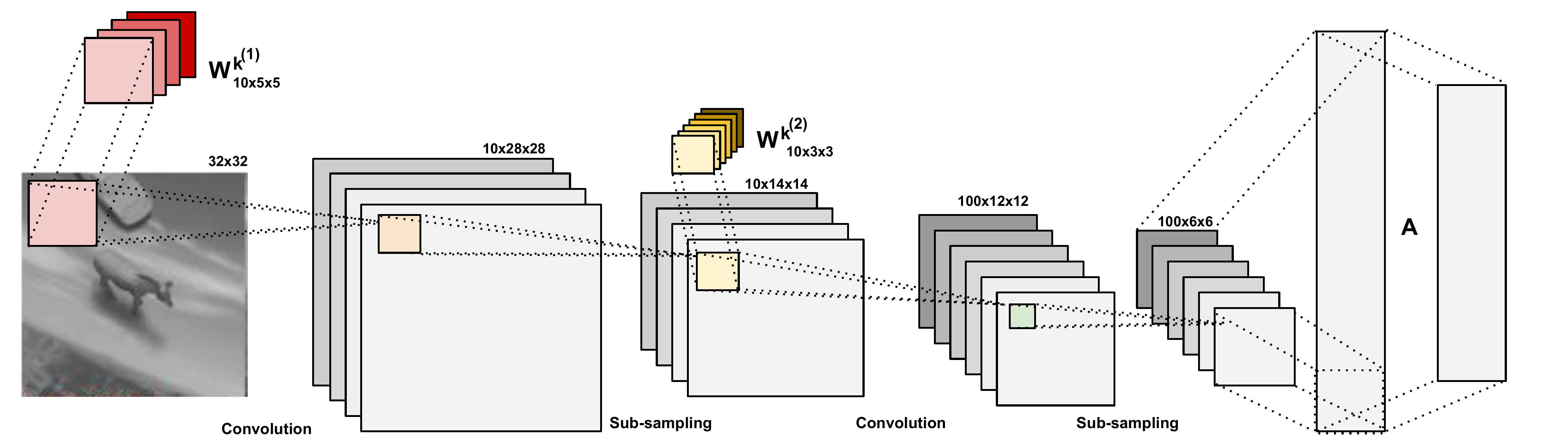}
    \end{minipage}%
\caption{Convolutional CCML with 2 convolution and subsampling layers. This architecture was used in our
  experiments.}
\label{fig:cccml}
\end{figure*}
\subsection{Towards a local search}
%
For each query point, Eq.~\ref{eqn:org_cne} implies a nearest neighbour search in each
class. However, a query may have nearest neighbours in some
classes that are globally very far away. Intuitively, it
does not make sense to waste modeling power in repelling them
further. We can capture this intuition while learning the distance
metric by only considering the probability of a single event: the
probability that a point is assigned to its true class. In other
words, we do not need to explicitly formulate the probability of
assignment to each class. We define the probability of
``correct-class'' assignment to be
%
%
\begin{multline}
p_{i}^{C_i} = \exp\left(-\frac{1}{k}\sum_{j=1}^k || Ax_i - \text{NN}^{C_i}_j(Ax_i) ||^2\right) \\\
\Bigg[\exp\left(-\frac{1}{k}\sum_{j=1}^k || Ax_i -
             \text{NN}^{C_i}_j(Ax_i) ||^2\right) + \\
            \exp\left(-\frac{1}{k}\sum_{j=1}^k || Ax_i -
        \text{NN}^{\bar{C_i}}_j(Ax_i) ||^2\right)\Bigg]^{-1}~,\
    \label{eqn:new_cne}
\end{multline}

\noindent where $\text{NN}^{\bar{C_i}}_j(Ax_i)$ means that neighbours
of $Ax_i$ are chosen from all classes \emph{except} $C_i$. We still
achieve the same effect if we replace Eq.~\ref{eqn:org_cne} by
Eq.~\ref{eqn:new_cne} in Eq.~\ref{eqn:expectation}, where neighbouring
points of the same class are pulled together, and neighbouring
points in different classes are repelled. Note that we cannot use this
probability as part of a decision rule, since class labels
are not available at test time. However, in practice, we use it in the objective
for learning a distance metric and apply either Eq.~\ref{eqn:ccknn}
or the simple KNN rule in the projected space at test time. Of course,
we expect (and verify in the experiments) that CCKNN improves
performance.

\begin{algorithm}[ht]
    \caption{Class Conditional K-Nearest Neighbour (CCKNN) test-time decision rule.}
    \label{algo:ccknn}
    \begin{algorithmic}
        \STATE Partition the data by classes: $x^{C_1}, x^{C_2}, ..., x^{C_n}$
        where $n$ is the number of classes.
        \STATE Compute the class prior for $C_i$ based on its frequency in the dataset $\mathcal{D}$:
            $p(C_i) = \frac{\text{number of $x_i$ belonging to $C_i$}}{\sum_j \text{number of $x_j$ belonging to $C_j$}}$
        \FOR {each class $C_i$}
            \STATE Compute the distance between $X_{test}$ and $X^{C_i}$
            \STATE Find the $k$-nearest distances - and store in $D_i$
        \ENDFOR
        \STATE Compute the variance of the distances, $\mathbf{\sigma} = \mathrm{Var}[[D_1;D_2;\cdots;D_c]]$
        \FOR {each class $C_i$}
            \STATE Compute the likelihood under a Gaussian distribution,
                $p(x_{test}|C_i) = \mathcal{N}(D_{ij} | 0, \mathbf{\sigma})$ $\forall j=1,...,k$
            \STATE Compute the score, which is proportional to the posterior probability, $                    p(C_i|x_{test}) \propto p(x_{test}|C_i) p(C_i)$
        \ENDFOR
    \end{algorithmic}
\end{algorithm}

Although both variants of CCML are bounded by the $O(N^2)$ task of
computing distances in projected space, we find that practically,
sorting distances twice rather than per-class gives a modest
speed-up. In addition, the burden of computing pairwise distances can
be mitigated by employing an approximation method such as KD-trees or
the AESA algorithm \cite{aesa}. More important than the modest gain in
speed, we find that this variant of CCML improves performance,
which is consistent with the intuition above.


\section{Extension to the Convolutional Setting}
Recently, convolutional neural networks have become much more popular
due to their strong performance on vision benchmarks. These architectures
naturally incorporate prior knowledge about the 2d structure of
images.  As in \cite{drlim}, we can construct a nonlinear variant of CCML by
replacing the purely linear transformation by several convolutional
layers, plus one or more fully-connected layers.  The $l^{th}$
convolutional and subsampling layer are formulated as

 \begin{align}
     z_j^{k^{(l)}} &= f\left(\sum_{j\in M_k} x_j^{(l-1)} * W^{k^{(l)}} + b_j^{k^{(l)}}\right)
     \label{eqn:conv}\\
     x_j^{k^{(l)}} &= \text{ down}(z_j^{(l)}) + c_j^{k^{(l)}}
     \label{eqn:subsam}
 \end{align}

\noindent respectively, where $*$ is the convolution operator, $M_j$
is the selection of inputs from the previous layer (``input maps''),
$f$ is an activation function, and $\lbrace W^k, b_j^k, c_j^k\rbrace$
are the parameters of the model.  The convolutional step (filtering)
is expressed in Eq.~\ref{eqn:conv}.  Typically, the rectified linear
(ReLU) or $\tanh$ function are used as the activation function
$f$. The sub-sampling layer is expressed in Eq.~\ref{eqn:subsam} and
we use max-pooling for the sub-sampling function $\text{down}(\cdot)$,
which selects the maximum value among the elements in each local $n_p
\times n_p$ region.  In practice, the bias $c_j^k$ is set to zero.

We can make a substitution in Eq.~\ref{eqn:new_cne},
replacing the input $x_i$ by the flattened output of
the last convolutional and subsampling stage.
We use the same objective (Eq.~\ref{eqn:expectation}), backpropagating
through all the fully-connected and convolutional layers.  We denote
the convolutional CCML as ConvCCML. The 2-layer ConvCCML architecture
we used in our experiments is depicted in Fig.~\ref{fig:cccml}.
%
\section{Experiments}
\label{experiments}

We evaluated CCML\footnote{Code can be found at \url{https://github.com/jiwoongim/CCML}} based on qualitatively assessing the distance metric
by visualization using a synthetic dataset and quantitatively evaluating nearest neighbour
classification and retrieval across four real datasets described below. In
contrast to previous work on NBNN \cite{Boiman2008,mccann2012local}, we
avoid the use of computing descriptors. Instead, we compare NN or NBNN
algorithms on raw pixels to learned representations obtained by
distance metric learning.
%

\subsection{Qualitative evaluation on synthetic data}
\begin{figure}[htb]
\vspace{-6mm}
\centering
    \begin{minipage}[b]{1.0\linewidth}
    \includegraphics[width=1.0\textwidth]{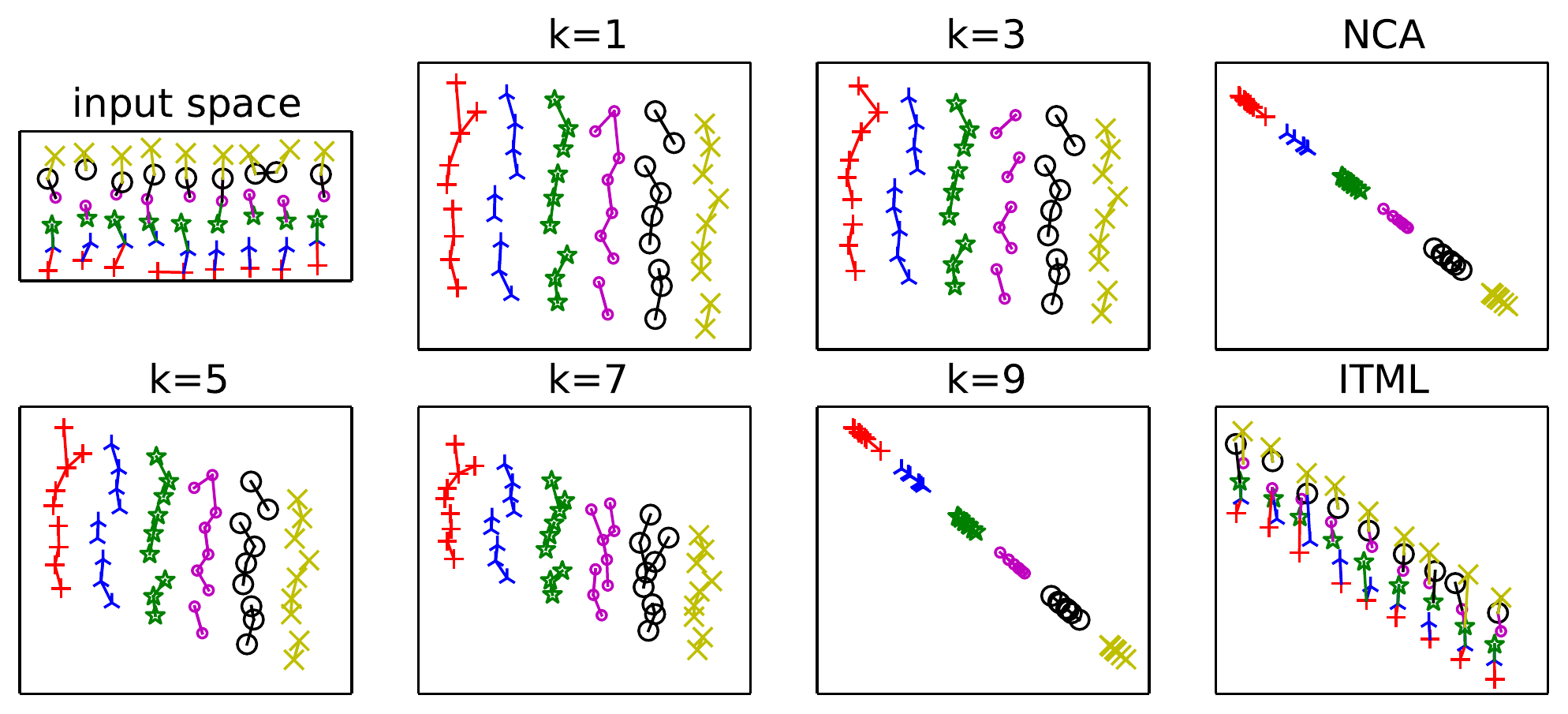}
    \end{minipage}%
\caption{CCML, NCA, and ITML embeddings of the Sandwich
  dataset. For CCML, various choices of $k$ are used. Nearest neighbours are shown as links. Colour and symbol
  indicate class.}
\label{fig:sandwich_fig}
\end{figure}

We used a synthetic dataset called ``Sandwich'' in order to
visualize the behaviour of CCML embedding, with respect to the choice of
$k$.  Fig.~\ref{fig:sandwich_fig} shows the embedding at various
settings of $k$, with nearest neighbour points as linked and
colours/symbols indicating class. In the input space, we can see that
every data point is linked with a data point in another class, which
indicates that the Euclidean metric is a poor choice for this
dataset. However, even after just using $k=1$ for learning the metric,
CCML alleviates the
problem. 
We see that for small values of $k$, data points are clustered locally
with other points from the same class. As we increase $k$, the
clustering becomes more global. This implies that, compared to NCA,
the scale of the clustering can be controlled with the choice of
$k$. This illustrates that our metric can be tuned to learn
a local or global distance metric depending on the data (e.g.~whether
the data is highly multi-modal or not).
The data points projected by NCA ($k=1$) and
Information-Theoretic Metric Learning (ITML) \cite{davis2007itml} are also shown.

\subsection{Datasets}
Here we will describe the four datasets that were considered in the
experiments described in the remainder of this section.

\noindent {\bf The UCI wine dataset} contains 178
examples. We took the standard approach (e.g.~as used by
\cite{weinberger2006lmnn}) of randomly separating the data into 70\%
training examples. 
The data consists of three different types of wines
with 13 attributes that describe each wine. Although the
dataset is small both in number of examples and dimensionality,
we included it because of its common use in the distance metric
learning literature (c.f.~\cite{davis2007itml,lui2012lsml,qi2009itml,weinberger2006lmnn}).

\noindent {\bf The MNIST dataset} contains 60,000 training and 10,000 test
images that are 28 $\times$ 28 pixel images of handwritten digits from
the classes 0 to 9. 
From the 60,000 training examples, we used 10,000 as validation
examples to tune the hyper-parameters in our model.

\noindent {\bf The ISOLET dataset} contains 7,797 instances and 617
features. The task is to predict which letter-name was spoken from
various people, and the features include: spectral coefficients,
contour features, pre-sonorant features, and post-sornorant
features. From the 7,797 examples, the dataset is pre-split into 6,238
(80\%) training examples and 1,559 (20\%) test examples.

\noindent {\bf The small NORB dataset} contains 24,300 training examples and
24,300 test examples that are stereo pairs of 32 $\times$ 32
images. It consists of 50 toy objects from five different classes:
four-legged animals, humans, airplanes, trucks, and cars.
Objects are captured under different lighting conditions, elevation,
and azimuth to assess the invariance properties of classifiers.\looseness=-1

\setlength{\tabcolsep}{4pt}
\begin{table*}[htp]
  \caption{Classification error rates across datasets. After learning or selecting a metric,  we apply either
    a standard (KNN) or class-conditional (CCKNN) decision rule. For
    the Euclidean metric, we considered two feature representations:
    PCA retaining 99\% of the variance, and PCA with an optimal number of components selected on a
    validation set in parentheses. Note that the convolutional models
    were only applied to image-based datasets.}
\label{tab:classification}
\centering
\begin{tabular}{l|ll|ll|ll|ll}
    \bf{Metric} & \multicolumn{2}{c|}{\bf{WINE}} &\multicolumn{2}{c|}{\bf{ISOLET}} & \multicolumn{2}{c|}{\bf{MNIST}} & \multicolumn{2}{c}{\bf{NORB}}\\
        \indent & KNN & CCKNN& KNN & CCKNN& KNN & CCKNN& KNN & CCKNN \\ \hline
    Euclidean (PCA w/ 99\% variance) & 3.99 & 3.29 & 8.79  & 7.12    & 2.91 & 2.71       & 22.13 & 22.13     \\ 
    Euclidean (PCA w/ optimal \# of comp.)  & 4.31 \small{(6)} & 3.67 \small{(6)} & 8.66 \small{(300)} & 6.80 \small{(200)} & 2.48 \small{(20)} & 2.19 \small{(10)} & 19.75 \small{(120)} & 19.40 \small{(120)}\\ 
    ITML \cite{davis2007itml}    & 4.44 & 3.71 & 9.49 & 7.45 & 2.93 & 2.75 & 20.73  & 20.53   \\
    NCMC   \cite{Mensink2013}    & 3.51 & 2.96 & 5.13 & 5.06 & 3.21 & 3.09 & 20.87  & 20.34    \\ 
    LMNN\cite{weinberger2006lmnn}& 2.59 & 2.59 & 5.33 & 5.38 & 2.11 & 2.1    & 15.14  & 14.61   \\
    NCA   \cite{Goldberger}      & 2.74 & 2.60 & 4.24 & 4.05 & 2.65 & 2.45 & 13.00  & 12.83    \\ 
    \hline
    CCML                         & 2.13 & \bf{2.04} & 3.66 & \bf{3.60} & 1.91 & 1.77 & 12.15 & 11.89    \\ 
    \hline\hline
    ConvNet (1-layer)            &  \multicolumn{4}{|c|}{\multirow{4}*{ }}  & \multicolumn{2}{|c|}{1.13}  & \multicolumn{2}{|c}{9.25}     \\ 
    ConvNet (2-layer)            &  \multicolumn{4}{|c|}{}  & \multicolumn{2}{|c|}{1.12}  & \multicolumn{2}{|c}{7.96}    \\ \cline{1-1}\cline{6-9}      
    ConvCCML (1-layer)            & \multicolumn{4}{|c|}{}  & 1.36 & 1.30 & 6.84  & 6.80     \\ 
    ConvCCML (2-layer)            & \multicolumn{4}{|c|}{}  & 0.91 & \bf{0.83} & \bf{6.59}  & 6.65    \\ 
\end{tabular}
\vspace{-5mm}
\end{table*}
\begin{figure*}[ht]
\centering
    \begin{minipage}[b]{.33\linewidth}
    \includegraphics[trim=.8cm 0cm 1.75cm 0cm, clip=true, width=1.0\textwidth]{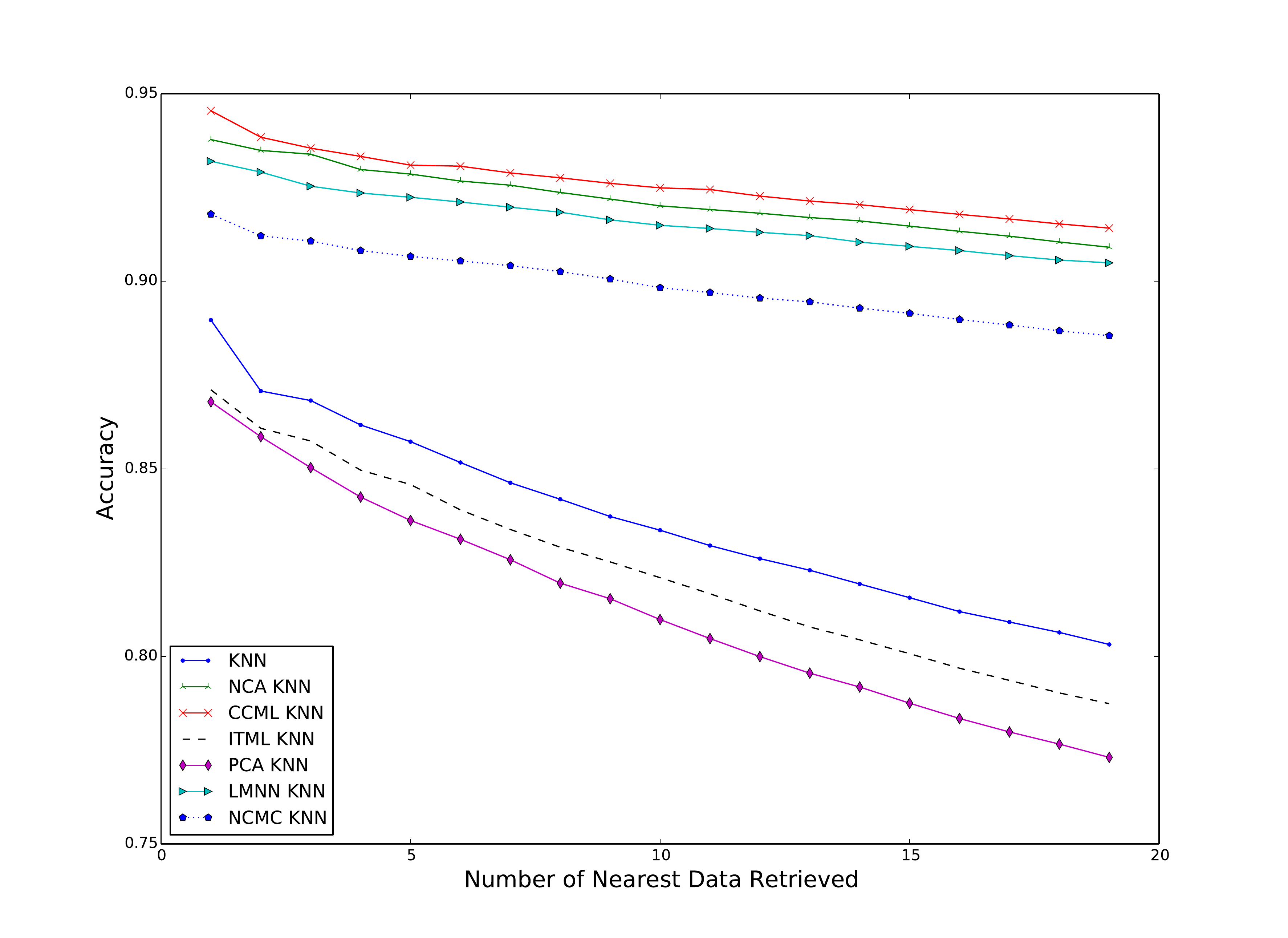}
    \subcaption{ISOLET}
    \label{fig:retrival_isolet}
    \end{minipage}%
    \begin{minipage}[b]{.33\linewidth}
    \includegraphics[trim=.8cm 0cm 1.75cm 0cm, clip=true, width=1.0\textwidth]{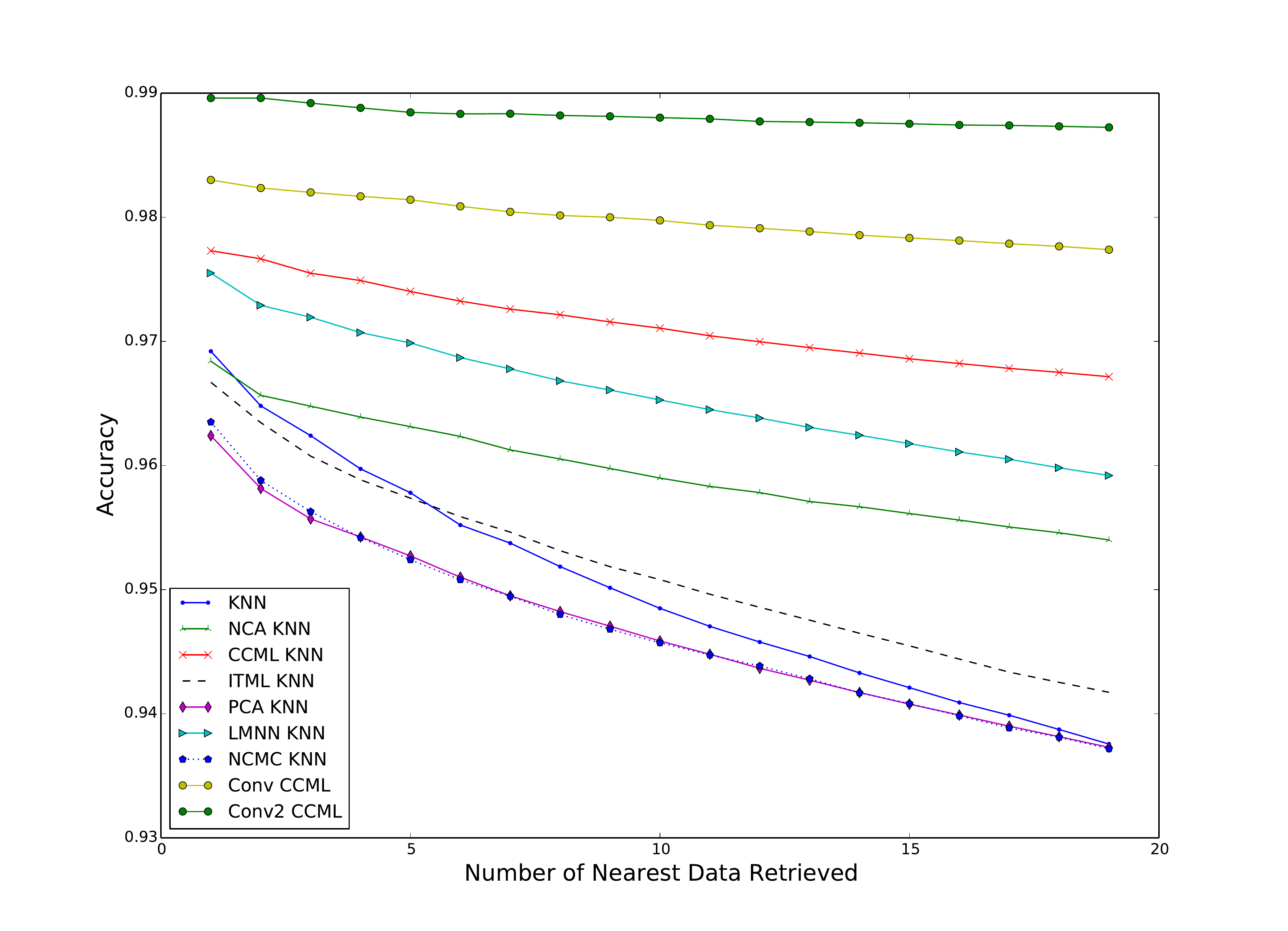}
    \subcaption{MNIST}
    \label{fig:retrival_norb}
    \end{minipage}
    \begin{minipage}[b]{.33\linewidth}
    \includegraphics[trim=.8cm 0cm 1.75cm 0cm, clip=true, width=1.0\textwidth]{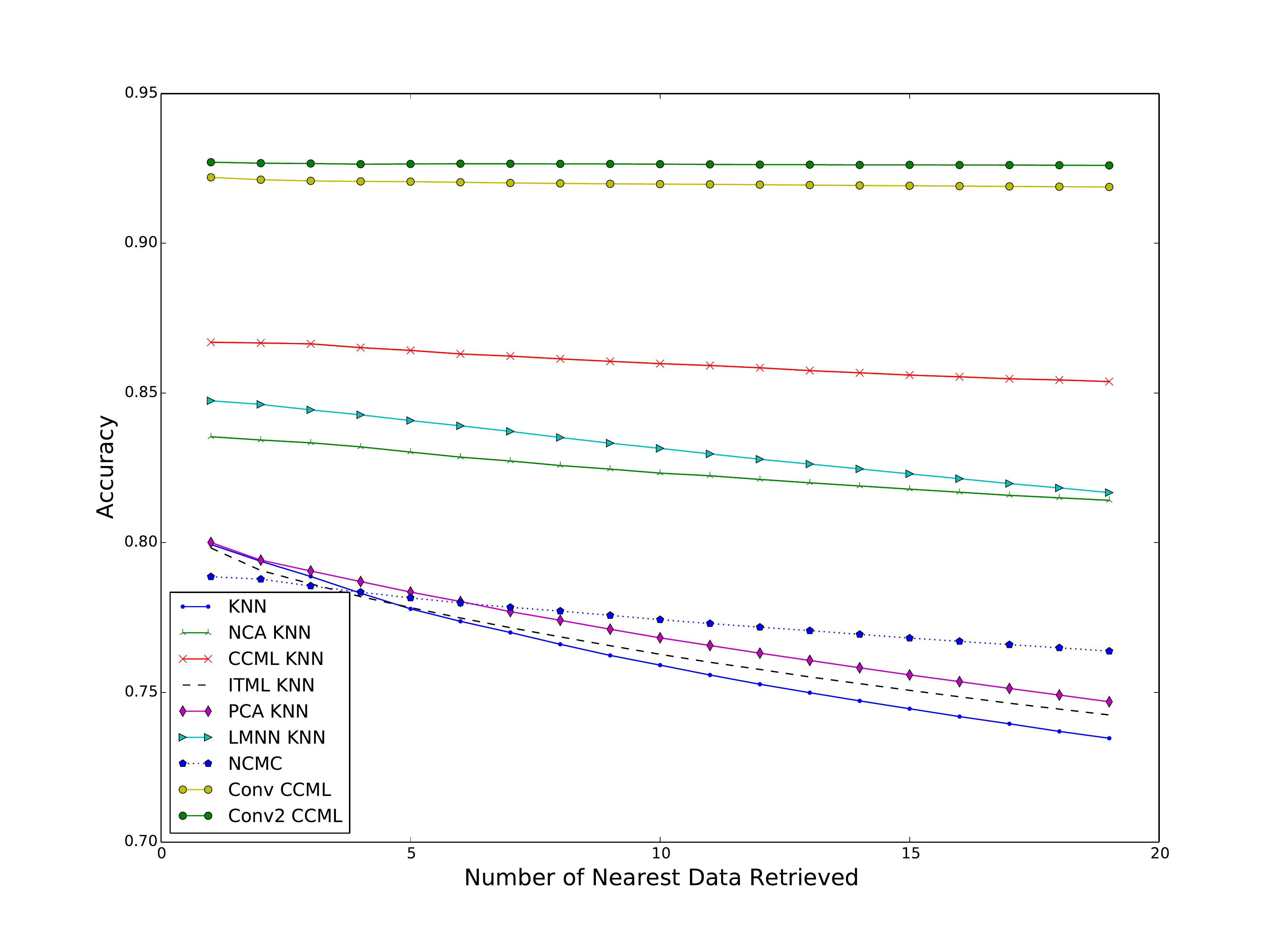}
    \subcaption{NORB}
    \label{fig:retrival_norb}
    \end{minipage}%
\caption{Retrieval quality w.r.t.~the number of images
  retrieved using KNN on the ISOLET, MNIST, and NORB datasets.}
\label{fig:retrivals}
\vspace{-5mm}
\end{figure*}
\subsection{Classification Performance}
We now compare several metric learning techniques based on nearest
neighbour classification. With the exception of the convolutional
models, we applied principal component analysis (PCA) before training
to reduce the dimensionality of the data, retaining 99\% of the
variance. For the NORB dataset, we preprocessed with Local Contrast
Normalization \cite{Pinto2008}. For all other datasets, we used raw
pixels as input. We used the usual training and test splits for
ISOLET, MNIST, and NORB. For the wine dataset, we ran 10-fold
cross-validation to obtain the results. Hyperparameters $k$, learning
rate, weight-decay cost, batch size, and size of the projected
dimension were selected from discrete ranges and chosen based on a
held-out validation set.

For classification, we measured accuracy using Euclidean
distance on the original space and transformed space via PCA, NCA,
ITML \cite{davis2007itml}, NCMC \cite{Mensink2013}, LMNN
\cite{weinberger2006lmnn}, and CCML. Results are shown in Table
\ref{tab:classification} and more details on each of the baselines are given
in Section \ref{related}.  We applied both KNN and CCKNN to each
metric, in an attempt to separate the gain from the type of NN
decision rule to the gain from the particular distance metric used.
The results illustrate that employing the CCKNN classifier, rather than
KNN, consistently achieves better performance.  Moreover, the distance
metric learned by NCA improves accuracy, but applying CCML improves
accuracy by a bigger margin. As expected, the learned CCML metric
combined with a CCKNN decision rule achieves the best performance. Our
method outperforms NCA, ITML, NCMC, and LMNN\footnote{According to
  \cite{weinberger2006lmnn}, the MNIST classification error on LMNN is
  1.31\%. We were unable to reproduce these results, even using the
  authors' code.} across all datasets.

For the image-based datasets, we evaluated a
convolutional version of our model, comparing it to an identical
convolutional architecture trained discriminatively (denoted ConvNet).
The 1 and 2-convolutional layer variants of CCML are denoted by
``convCCML(1-layer)'' and ``convCCML(2-layer)'' in
Table~\ref{tab:classification}.  
The simplest architecture, convCCML(1-layer) used 10 5 $\times$ 5
filters. The other architecture, convCCML(2-layer), used 10 filters
for each layer with 5 $\times$ 5 filters on the first convolutional
layer and 3 $\times$ 3 filters on the second layer. The pooling size
was 2 by 2 for all convolutional architectures.
%
ConvCCML significantly outperformed the baseline
ConvNet, as well as the vector-based metric learning techniques.
%

%
\subsection{Data retrieval}
\begin{figure*}[t]
\centering
    \begin{minipage}[b]{1.0\linewidth}
    \includegraphics[width=1.0\textwidth]{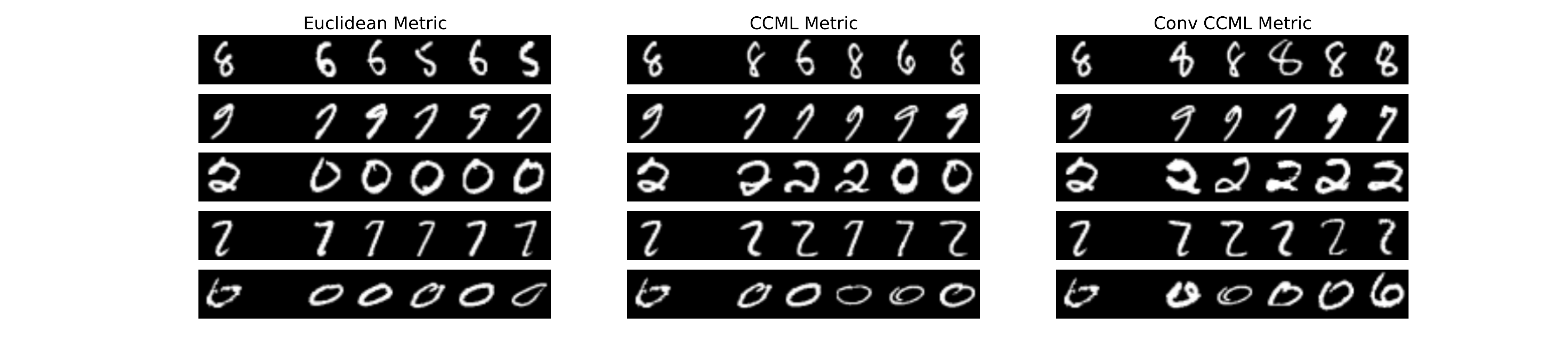}
    \subcaption{Retrieval on the MNIST dataset}\label{fig:mnist_retrieval}
    \end{minipage}
    \begin{minipage}[b]{1.0\linewidth}
    \includegraphics[width=1.0\textwidth]{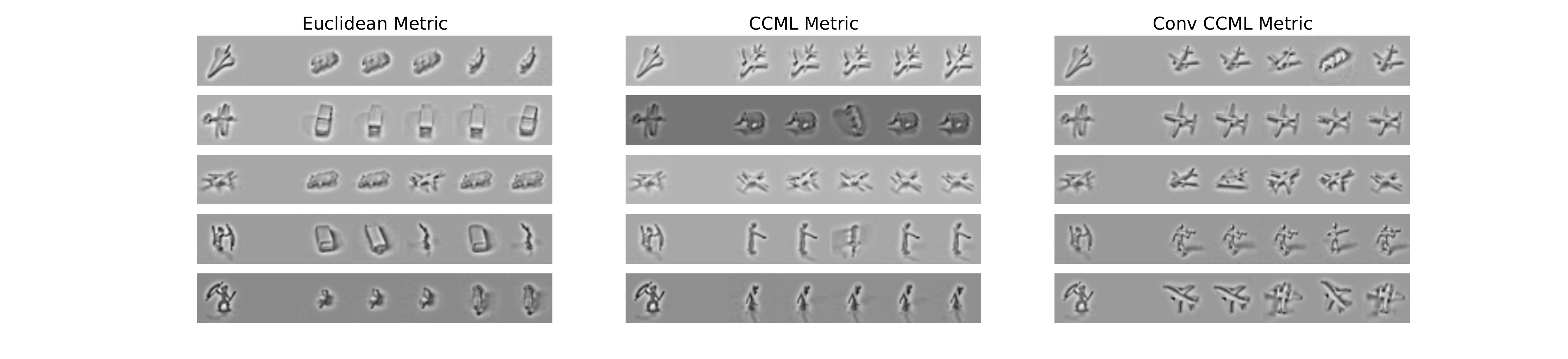}
    \subcaption{Retrieval on the NORB dataset}\label{fig:norb_retrieval}
    \end{minipage}
    \caption{Visualization of the retrieval task:
      \subref{fig:mnist_retrieval}) MNIST,
      \subref{fig:norb_retrieval}) NORB.
      For each subplot, the first column shows the query image. Columns
      2-6 show the five data points that were closest to the target
      image as per the respective metric. The last row of each
      subplot is an example of a failure mode under each method.}
    \label{fig:retrieval_data}
    \vspace{-3mm}
\end{figure*}
Distance metric learning is frequently used for the retrieval of
relevant examples, such as images or documents. In this section,
we explore the performance of various distance metric learning methods
using a KNN rule for retrieval.
Retrieval quality was measured using normalized Discounted Cumulative Gain (nDCG), where
$\text{DCG}_k = \sum^k_{i=1} \frac{2^{rel_i}-1}{\log_2{(i+1)}}$, $\text{nDCG}_k = \frac{\text{DCG}_k}{\text{IDCG}_k}$
and $\text{IDCG}_k$ is
the maximum possible (ideal) DCG for a given set of queries.
nDCG, commonly used in information retrieval,  measures the performance of retrieved data quality based on the graded relevance
of the retrieved entities. We use a binary indicator of class match as a
relevance score. This task focuses more on evaluating
the individual returned examples from the database, rather than their
aggregate vote. This is a more explicit form of
evaluating the ability of the learned distance metric to capture
perceptual similarity than classification.\looseness=-1

For each of the datasets (ISOLET, MNIST, and NORB), we returned the $k$
nearest neighbours of each query image and reported the fraction that
were from the true class (Fig.~\ref{fig:retrivals}).
%
%
%
Note that CCKNN cannot be employed in the retrieval task since CCKNN
compares each item to a class, returning a class decision. Here, we
want to compare the query example to each individual item in the
database and return the most relevant item.  The results on all three
datasets illustrate that retrieval using a distance metric
significantly outperforms retrieval using a Euclidean distance, and it
consistently shows a large relative gain between CCML and other
methods like NCA or LMNN.
We also evaluated the retrieval quality of convolutional variants
of CCML.
The more flexible convCCML was shown to significantly outperform the other
methods.

Fig.~\ref{fig:retrieval_data} visualizes retrieval and failure modes. Each
row shows the five nearest neighbours
to the query using the Euclidean,
CCML and convCCML distance metrics.
Both MNIST and NORB target samples in the first four rows of Fig.~\ref{fig:retrieval_data}
were incorrectly classified by KNN, but correctly classified by KNN after
being transformed by CCML.
The first, second, third, and fourth rows of
Fig.~\ref{fig:mnist_retrieval} are dominated by images of `sixes',
`sevens', `zeros', and `sevens', even though the query images were `eight',
`nine', `two', and `two'. Although performance improves with CCML, it
still produces a few errors: not all of the five nearest neighbours
 share a class with the query. Indeed, this is amended by using
convCCML, where all five nearest neighbours are correct.

We also show visual retrieval performance on the NORB dataset
(Fig.~\ref{fig:norb_retrieval}).
Discrimination among the five different categories of 3D objects is
more difficult than with MNIST, because images from the same
category can significantly differ in appearance. This is reflected
in their pixel-space distances.
The last row in both the MNIST and NORB retrieval examples show
failure modes of convCCML.
Even though the retrieved images are incorrectly classified,
we can see that the images retrieved using the convCCML distance measure do capture some sort of similarity in terms of
geometry.
Qualitatively, among the misclassified examples,
images retrieved using convCCML appear to be more sensible.

\section{Related Work}
\label{related}
Many metric learning methods, particularly $k$-NN, have been proposed
for non-parametric classification. A large
number of these methods attempt to learn a Mahalanobis distance. A
well-known example is Neighbourhood Components Analysis (NCA), which
learns an optimal Mahanalobis distance for 1-NN \cite{Goldberger}.  It
maximizes a stochastic variant of the leave-one-out criterion, and it
is trained through stochastic gradient descent. Since our work is
inspired by NCA, we have already discussed its formulation in
Section~\ref{sec:NCA}.

Salakhutdinov and Hinton extended NCA to the non-linear setting by incorporating a multi-layer
neural network into the embedding and training it with backpropagation
\cite{Salakhutdinov2007}. They incorporated unsupervised pre-training \cite{Hinton2006}
into their formulation and regularized with an auto-encoder
objective. In a similar manner, we demonstrate the extension of our
method to the non-linear setting by utilizing an embedding based on
convolutional neural networks.

One drawback of the NCA criterion is that it only finds the best
metric for a 1-NN classifier since it stochastically selects a single
neighbour per point. Tarlow et al.~address this issue by extending to
stochastic $k$-neighbourhood selection
\cite{tarlow2013stochastic}. Though this method is theoretically
sound, in practice, optimization is complicated. Moreover, the technique is
memory-intensive in the case of large datasets, and as a result, it is
not as straightforward to extend to non-linear mappings.

Another well-known metric learning algorithm is Large Margin Nearest
Neighbour (LMNN) \cite{weinberger2006lmnn}, which also learns a
Mahalanobis distance for $k$-NN by maximizing a hinge loss
function. The optimization is formulated as a semi-definite
programming problem, which is convex. However, it
does not permit non-linear embeddings as in
\cite{Salakhutdinov2007} or the present work.

Mensink \etal~proposed a non-parametric classification method called
the nearest class mean classifier (NCMC).  Superficially, this
algorithm may appear similar to ours, but operationally, it is
not. NCMC is a prototype-based algorithm that compares the distance
between a point to the mean over all points from each class. On the
other hand, our method discriminates classes based on computing the
average of $k$-local distances from each class. Optimizing for NCMC
and CCKNN is not the same (see Section \ref{sec:ccknn}).  An advantage
of our method is its local to global scaling (as shown in
Fig.~\ref{fig:sandwich_fig}). Moreover, NCMC learns from image
descriptors such as SIFT, whereas our proposed metric learns features
from pixels.

Other types of metric learning were also explored in the learning
literature, including Information-Theoretic Metric Learning
\cite[ITML]{davis2007itml}.  ITML minimizes the log determinant
divergence subject to linear constraints based on Bregman
optimization, which is convex like LMNN.  Also, other variants of
ITML that use Bregman optimization exist, such as \cite{Wu2011,
  Banerjee2005} or generalized maximum entropy \cite{Yang2010,
  Dudik2006}. Though these methods have compelling theoretical
results, none have emerged as dominant in practice.

\section{Conclusion}

Inspired by Neighbourhood Components Analysis which optimizes a
distance metric for 1-nearest neighbour classification, we show how to
learn a distance metric optimized for class-conditional nearest
neighbour.  Its benefit is twofold: i) we show improved performance
over existing metric learning methods, including ITML, NCMC, LMNN, and
NCA; and ii) by removing the dependency on engineered descriptors, we
extend the method's applicability to non-image domains.

We demonstrated linear and nonlinear versions of our technique,
the latter based on convolutional neural networks. For non-signal type
data, i.e.~data inappropriate for convnets, any nonlinear
function can be used, as long as it is differentiable. CCML is
trained by gradient descent (applying backpropagation in the
case of multi-layer mappings) and it is suitable for batch or online
learning. Moreover, it can be adjusted to be more local or global by
tuning a single parameter.

Due to the nature of our decision rule, our metric learning approach is
limited to datasets which have pre-defined class labels. We intend to
investigate whether the approach can be extended to weaker forms of
labeling.




%
\bibliographystyle{IEEEtran}
\bibliography{ccml}

\begin{thebibliography}{10}
\providecommand{\url}[1]{#1}
\csname url@samestyle\endcsname
\providecommand{\newblock}{\relax}
\providecommand{\bibinfo}[2]{#2}
\providecommand{\BIBentrySTDinterwordspacing}{\spaceskip=0pt\relax}
\providecommand{\BIBentryALTinterwordstretchfactor}{4}
\providecommand{\BIBentryALTinterwordspacing}{\spaceskip=\fontdimen2\font plus
\BIBentryALTinterwordstretchfactor\fontdimen3\font minus
  \fontdimen4\font\relax}
\providecommand{\BIBforeignlanguage}[2]{{%
\expandafter\ifx\csname l@#1\endcsname\relax
\typeout{** WARNING: IEEEtran.bst: No hyphenation pattern has been}%
\typeout{** loaded for the language `#1'. Using the pattern for}%
\typeout{** the default language instead.}%
\else
\language=\csname l@#1\endcsname
\fi
#2}}
\providecommand{\BIBdecl}{\relax}
\BIBdecl

\bibitem{Csurka04visualcategorization}
G.~Csurka, C.~R. Dance, L.~Fan, J.~Willamowski, and C.~Bray, ``Visual
  categorization with bags of keypoints,'' in \emph{ECCV Workshop on
  Statistical Learning in Computer Vision}, 2004.

\bibitem{lazebnik2006beyond}
S.~Lazebnik, C.~Schmid, and J.~Ponce, ``Beyond bags of features: Spatial
  pyramid matching for recognizing natural scene categories,'' in
  \emph{Proceedings of CVPR}, 2006.

\bibitem{Krizhevsky:2012}
A.~Krizhevsky, I.~Sutskever, and G.~E. Hinton, ``Imagenet classification with
  deep convolutional neural networks,'' in \emph{Proceedings of NIPS}, 2012.

\bibitem{szegedy2014going}
C.~Szegedy, W.~Liu, Y.~Jia, P.~Sermanet, S.~Reed, D.~Anguelov, D.~Erhan,
  V.~Vanhoucke, and A.~Rabinovich, ``Going deeper with convolutions,''
  \emph{arXiv preprint arXiv:1409.4842}, 2014.

\bibitem{Boiman2008}
O.~Boiman, E.~Shechtman, and M.~Irani, ``In defense of nearest-neighbor based
  image classification,'' in \emph{Proceedings of CVPR}, 2008.

\bibitem{mccann2012local}
S.~McCann and D.~G. Lowe, ``Local naive {Bayes} nearest neighbor for image
  classification,'' in \emph{Proceedings of CVPR}, 2012.

\bibitem{lowe2004distinctive}
D.~G. Lowe, ``Distinctive image features from scale-invariant keypoints,''
  \emph{IJCV}, vol.~60, no.~2, pp. 91--110, 2004.

\bibitem{mori2005efficient}
G.~Mori, S.~Belongie, and J.~Malik, ``Efficient shape matching using shape
  contexts,'' \emph{IEEE Tr.~PAMI}, vol.~27, no.~11, pp. 1832--1837, 2005.

\bibitem{behmo2010towards}
R.~Behmo, P.~Marcombes, A.~Dalalyan, and V.~Prinet, ``Towards optimal naive
  {Bayes} nearest neighbor,'' in \emph{Proceedings of ECCV}, 2010.

\bibitem{Weston2008}
J.~Weston, F.~Ratle, and R.~Collobert, ``Deep learning via semi-supervised
  embedding,'' in \emph{International Conference of Machine Learning (ICML)},
  2008.

\bibitem{Bian2014}
W.~Bian, T.~Zhou, M.~Martinez, Aleix, G.~Baciu, and D.~Tao, ``Minimizing
  nearest neighbor classification error for nonparametric dimension
  reduction,'' \emph{IEEE Transactions on Neural Networks and Learning
  systems}, pp. 1588--1594, 2014.

\bibitem{Xu2014}
C.~Xu, D.~Tao, C.~Xu, and Y.~Rui, ``Large-margin weakly supervised
  dimensionality reduction,'' in \emph{International Conference of Machine
  Learning (ICML)}, 2014.

\bibitem{Saul2003}
L.~K. Saul and S.~T. Roweis, ``Think globally, fit locally: Unsupervised
  learning of low dimensional manifolds,'' \emph{JMLR}, vol.~4, pp. 119--155,
  2003.

\bibitem{Hinton2002}
G.~Hinton and S.~Roweis, ``Stochastic neighbor embedding,'' in
  \emph{Proceedings of NIPS}, 2002.

\bibitem{Belkin2003}
M.~Belkin and P.~Niyogi, ``Laplacian eigenmaps for dimensionality reduction and
  data representation,'' \emph{Neural Computation}, vol.~15, no.~6, pp.
  1373--1396, 2003.

\bibitem{bengio2004out}
Y.~Bengio, J.-F. Paiement, P.~Vincent, O.~Delalleau, N.~Le~Roux, and M.~Ouimet,
  ``Out-of-sample extensions for {LLE}, isomap, {MDS}, eigenmaps, and spectral
  clustering,'' \emph{Proceedings of NIPS}, vol.~16, pp. 177--184, 2004.

\bibitem{carreira2007laplacian}
M.~A. Carreira-Perpin{\'a}n and Z.~Lu, ``The {Laplacian} eigenmaps latent
  variable model.'' \emph{JMLR}, vol.~2, pp. 59--66, 2007.

\bibitem{kim2009covariance}
M.~Kim and V.~Pavlovic, ``Covariance operator based dimensionality reduction
  with extension to semi-supervised settings,'' in \emph{Proceedings of
  AISTATS}, 2009, pp. 280--287.

\bibitem{Goldberger}
J.~Goldberger, S.~Roweis, G.~Hinton, and R.~Salakhutdinov, ``Neighbourhood
  component analysis,'' in \emph{Proceedings of NIPS}, 2004.

\bibitem{tarlow2013stochastic}
D.~Tarlow, K.~Swersky, L.~Charlin, I.~Sutskever, and R.~Zemel, ``Stochastic
  k-neighborhood selection for supervised and unsupervised learning,'' in
  \emph{Proceedings of ICML}, 2013.

\bibitem{Salakhutdinov2007}
R.~Salakhutdinov and G.~Hinton, ``Learning a nonlinear embedding by preserving
  class neighbourhood structure,'' in \emph{Proceedings of AISTATS}, 2007.

\bibitem{bottou1998}
L.~Bottou, ``Online algorithms and stochastic approximations,'' in \emph{Online
  Learning and Neural Networks}.\hskip 1em plus 0.5em minus 0.4em\relax
  Cambridge University Press, 1998.

\bibitem{aesa}
J.~M. Vilar, ``Reducing the overhead of the {AESA} metric-space nearest
  neighbour searching algorithm,'' \emph{Information Processing Letters},
  vol.~55, no.~5, pp. 265--271, 1995.

\bibitem{drlim}
R.~Hadsell, S.~Chopra, and Y.~LeCun, ``Dimensionality reducton by learning an
  invariant mapping,'' in \emph{Proceedings of CVPR}, 2006.

\bibitem{davis2007itml}
J.~V. Davis, B.~Kulis, P.~Jain, S.~Sra, and I.~S. Dhillon,
  ``Information-theoretic metric learning,'' in \emph{Proceedings of ICML},
  2007.

\bibitem{weinberger2006lmnn}
K.~Q. Weinberger and L.~K. Saul, ``Distance metric learning for large margin
  nearest neighbor classification,'' \emph{JMLR}, vol.~10, no.~2, pp. 207--244,
  2009.

\bibitem{lui2012lsml}
E.~Y. Liu, Z.~Guo, X.~Zhang, V.~Jojic, and W.~Wang, ``Metric learning from
  relative comparisons by minimizing squared residual.'' in \emph{Proceedings
  of ICDM}, 2012, pp. 978--983.

\bibitem{qi2009itml}
G.-J. Qi, J.~Tanga, Z.-J. Zha, T.-S. Chua, and H.-J. Zhang, ``An efficient
  sparse metric learning in high-dimensional space via l1-penalized
  log-determinant regularization,'' in \emph{Proceedings of ICML}, 2009.

\bibitem{Mensink2013}
T.~Mensink, J.~Verbeek, F.~Perronnin, and G.~Csurka, ``Distance-based image
  classification: Generalizing to new classes at near-zero cost,'' \emph{IEEE
  Tr.~PAMI}, vol.~35, no.~11, pp. 2624--2637, 2013.

\bibitem{Pinto2008}
N.~Pinto, D.~D. Cox, and J.~J. DiCarlo, ``Why is real-world visual object
  recognition hard?'' \emph{PLoS Computational Biology}, vol.~4, no.~1, 2008.

\bibitem{Hinton2006}
G.~E. Hinton, S.~Osindero, and Y.-W. Teh, ``A fast learning algorithm for deep
  belief nets,'' \emph{Neural Computations}, vol.~18, pp. 1527--1554, 2006.

\bibitem{Wu2011}
L.~Wu, S.~Hoi, R.~Jin, and N.~Yu, ``Learning {Bregman} distance functions for
  semi-supervised clustering,'' \emph{IEEE Transactions on Knowledge and Data
  Engineering}, vol.~24, no.~3, pp. 478--491, 2011.

\bibitem{Banerjee2005}
A.~Banerjee, ``Clustering with {Bregman} divergences,'' \emph{JMLR}, vol.~6,
  pp. 1705--1749, 2005.

\bibitem{Yang2010}
T.~Yang, R.~Jin, and A.~K. Jain, ``Learning from noisy side information by
  generalized maximum entropy model,'' in \emph{Proceedings of ICML}, 2010.

\bibitem{Dudik2006}
M.~Dudik and R.~E. Schapire, ``Maximum entropy distribution estimation with
  generalized regularization,'' in \emph{Proceedings of COLT}, 2006.

\end{thebibliography}

\end{document}